\documentclass[11pt,a4paper]{article}
\usepackage[hyperref]{emnlp-ijcnlp-2019}

\usepackage{booktabs}       %
\usepackage{amsfonts}       %
\usepackage{nicefrac}       %
\usepackage{microtype}      %

\usepackage{amsmath}

\usepackage{times}

\usepackage{mathtools}
\usepackage{latexsym}
\usepackage{graphicx,xcolor}
\usepackage{multirow}
\usepackage{xspace}
\usepackage{tabularx}
\usepackage{arydshln}

\usepackage{stmaryrd}

\DeclareMathAlphabet{\mathcal}{OMS}{zplm}{m}{n}

\newcommand{\blockSize}{m}

\let\cconv*
\let\ccorr\star

\newcommand{\TransE}{{\text{TransE}}}
\newcommand{\HolE}{{\text{HolE}}}
\newcommand{\DistMult}{{\text{DistMult}}}
\newcommand{\ComplEx}{{\text{ComplEx}}}
\newcommand{\RESCAL}{{\text{RESCAL}}}
\newcommand{\BlockHolE}{{\text{BlockHolE}}}

\newcommand{\Cset}{\mathbb{C}}
\newcommand{\Rset}{\mathbb{R}}
\newcommand{\mat}[1]{\boldsymbol{\mathbf{#1}}}
\newcommand{\RE}{\mathop{\text{Re}}}

\newcommand{\FourierMatrix}{\mat{F}}

\newcommand{\transpose}{^{\mathrm{T}}}
\newcommand{\inv}{^{\mathrm{-1}}}
\newcommand{\diag}{\mathop{\text{diag}}}
\newcommand{\cir}{\mathop{\text{circ}}}

\newsavebox\tempbox

\usepackage{xcolor}

\usepackage{pifont}

\usepackage{tikz}
\usetikzlibrary{shapes,positioning,calc,
arrows,automata,intersections,matrix,decorations}

\aclfinalcopy %

\title{A Non-commutative Bilinear Model for Answering Path Queries in Knowledge Graphs}

\author{Katsuhiko Hayashi$\,{}^{\dagger,*}$ \\
  {\small\tt katsuhiko-h@sanken.osaka-u.ac.jp} \\\And
  Masashi Shimbo$\,{}^{\ddagger,*}$ \\
  {\small\tt shimbo@is.naist.jp}\\
  \AND
  \mdseries
  ${}^\dagger$Osaka University\\
  Suita, Osaka, Japan\\
  \And
  \mdseries
  ${}^\ddagger$NAIST\\
  Ikoma, Nara, Japan\\
  \And
  \mdseries
  ${}^*$Riken AIP\\
  Chuo-ku, Tokyo, Japan
\\}

\date{}

\begin{document}
\maketitle

\begin{abstract}
Bilinear diagonal models for knowledge graph embedding (KGE),
such as DistMult and ComplEx,
balance expressiveness and computational efficiency
by representing relations as diagonal matrices.
Although they perform well in predicting atomic relations,
composite relations (relation paths) cannot be modeled naturally by
the product of relation matrices,
as the product of diagonal matrices is commutative and hence invariant with the order of relations.
In this paper,
we propose a new bilinear KGE model, called BlockHolE,
based on block circulant matrices.
In BlockHolE,
relation matrices can be non-commutative,
allowing composite relations to be modeled by matrix product.
The model is parameterized in a way that covers a spectrum ranging from diagonal to full relation matrices.
A fast computation technique is developed on the basis of the duality of the Fourier transform of circulant matrices.
\end{abstract}

\section{Introduction}
\label{sec:intro}

Large-scale knowledge graphs \cite{survey} %
are indispensable resources for knowledge-intensive applications such as question answering, dialog systems,
and distantly supervised relation extraction.
A knowledge graph is a collection of
\emph{triplets} $(s,r,o)$
representing the fact that (binary) relation $r$ holds between subject entity $s$ and object entity $o$.
Although efforts continue to enrich existing knowledge graphs with more facts,
many facts are still missing~\cite{survey}.
Knowledge graph completion (KGC) aims to automatically detect missing facts in an \emph{incomplete} knowledge graph,
and has become an active field of research in recent years.

\begin{figure}[t]
\centering
\colorlet{myBlue}{blue!75!cyan}
\colorlet{myRed}{red!75!orange}

\begin{tabular}{l}
  \begin{tikzpicture}[xscale=2.35,yscale=2.5,baseline=-1.55,>=latex]
\tikzstyle{every state}=[minimum size=2mm]

\foreach \r/\th/\name/\dispname/\pos in {%
  2/33+9/will/William/south,
  2/33-9/harry/Harry/west,
  1/33/charles/Charles/east,
  0.9/11/x2/\ldots/west,
  0/0/elizabeth/Elizabeth/east,
  1/-11/and/Andrew/north,
  2/-11+9/x3/Beatrice/south,
  2/-11-9/x4/Eugenie/north,
  1/-33/x5/\ldots/west%
}{
  \path (\th-0:\r) node[state] (\name) {} node[outer sep=1.5mm, anchor=\pos, font=\scriptsize] (\name_label) {\dispname};
}

\begin{scope}
\node [anchor=south east] at (elizabeth_label.north west) {(a)};
\end{scope}

\begin{scope}
\tikzstyle{every node}=[sloped,fill=white,font=\scriptsize, inner sep=1pt]

\path[->] (elizabeth) edge node {motherOf} (charles)
                   edge node {motherOf} (x2) 
                   edge node {motherOf} (and)
		   edge node {motherOf} (x5);

\path[<-,bend left=30] (charles) edge[myBlue, thick] node {fatherOf$^{-1}$} (will);

\path[->] (charles) edge[myRed, thick] node {fatherOf} (will)
		    edge node {fatherOf} (harry);

\path[->] (and) edge[myBlue, thick] node {fatherOf} (x3)
		edge node {fatherOf} (x4);

\path[->,bend left=69] (charles) edge[myBlue, thick] node[pos=0.5] {brotherOf} (and);

\path[->,bend left=33] (will) edge[myRed, thick] node[pos=0.45] {brotherOf} (harry);

\path[<-,bend right=27] (charles) edge[myRed, thick] node[pos=0.55] {fatherOf$^{-1}$} (harry);

\end{scope}
\end{tikzpicture}
 \\
          \begin{tikzpicture}[xscale=2.1,yscale=2.2,baseline=-1.55,>=latex]
            \tikzstyle{every state}=[minimum size=2mm]

            \foreach \x/\y/\name/\dispname/\pos in {%
              0/0.0/actor1/William/below,
              1.0/0.0/char1//below,
              2.0/0.0/movie1//below,
              3.0/0.0/genre1//below%
            }{
             \path (\x,\y) node[state] (\name) {} node[\pos=.3em,font=\scriptsize] {\dispname};
            }
	    \begin{scope}
              \node at (-0.3, 0) {{(b)}};
	    \end{scope}
            \begin{scope}
              \tikzstyle{every node}=[fill=white,font=\scriptsize]
              \path[->] (actor1) edge[myBlue, thick] node {fatherOf$^{-1}$} (char1);

              \path[->] (char1) edge[myBlue, thick] node {brotherOf} (movie1) ;

              \path[->] (movie1) edge[myBlue, thick] node {fatherOf} (genre1);

            \end{scope}
          \end{tikzpicture}
 \\
\begin{tikzpicture}[xscale=2.1,yscale=2.2,baseline=-1.55,>=latex]
            \tikzstyle{every state}=[minimum size=2mm]

            \foreach \x/\y/\name/\dispname/\pos in {%
              0/0/actor2/William/below,
              1.0/0/char2//below,
              2.0/0/movie2//below,
              3.0/0/genre2//below%
            }{
              \path (\x,\y) node[state] (\name) {} node[\pos=.3em,font=\scriptsize] {\dispname};
            }
	    \begin{scope}
                \node at (-0.3,0) {{(c)}};
            \end{scope}
            \begin{scope}
              \tikzstyle{every node}=[fill=white,font=\scriptsize]

              \path[->] (actor2) edge[myRed, thick] node {brotherOf} (char2);

              \path[->] (char2) edge[myRed, thick] node {fatherOf$^{-1}$} (movie2) ;

              \path[->] (movie2) edge[myRed, thick] node {fatherOf} (genre2);

            \end{scope}

          \end{tikzpicture}

\end{tabular}
\caption{(a) A knowledge graph and (b,c) two relation paths starting from William.
}
\label{fig:kg}
\end{figure}
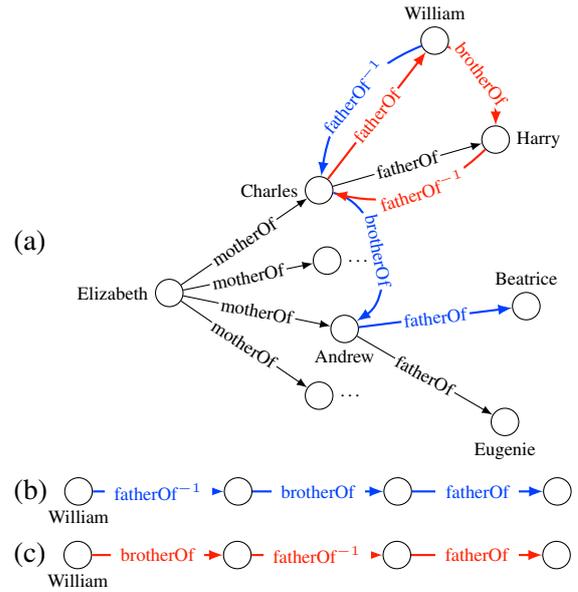

Knowledge graph embedding (KGE) is a promising approach to KGC.
It embeds entities and relations in vector space,
and defines a \emph{scoring function} $\phi(s,r,o)$ %
to evaluate the degree of factuality of a given triplet $(s,r,o)$
in terms of vector operations.

Bilinear KGE models are a popular choice for a scoring function,
along with those based on translation and neural networks.
RESCAL~\cite{rescal} adopts
a generic bilinear form as the scoring function, given by $\phi_\RESCAL(s,r,o)=\mat{e}_s\transpose\mat{R}_r\mat{e}_o$.
In this formula,
$ \mat{e}_s, \mat{e}_o $ 
are the $n$-dimensional vector embeddings of entities $s$ and $o$, respectively,
and $\mat{R}_r$ 
is the $n \times n$ matrix embedding of relation $r$.
Some of the more recent models have constrained the relation matrices to be diagonal. %
DistMult~\cite{distmult} and ComplEx~\cite{complex} are two such diagonal models.
HolE~\cite{hole} does not use diagonal relation matrices,
but has been shown \cite{eq} to be isomorphic to ComplEx.
These models have a smaller number of parameters than RESCAL, making them less prone to overfitting, and the performance is usually better.

While all these models were designed with a specific task of KGC in mind, i.e., computing the factuality of triplets,
another important task on knowledge graphs was pursued by \citet{guu} and \citet{path_reg}.
This latter task, called
\emph{path query answering (path QA)},
is to answer composite queries that consist of a cascade of relations,
as opposed to an atomic relation.
See Figure~\ref{fig:kg} for instance.
A query ``Is Beatrice a child of a paternal uncle of William?''
can be answered by predicting the truth value of the triplet
(William, fatherOf$^{-1}$/brotherOf/fatherOf, Beatrice)
where fatherOf$^{-1}$/brotherOf/fatherOf
is a binary relation not present in the knowledge graph as a relation (edge) label
but is composed of a cascade of three atomic relations.\footnote{%
  We regard inverse relations (e.g., $\text{fatherOf}^{-1}$) also as atomic relations.
}
Composite queries are also called \emph{path queries}, as they can be represented as paths in a knowledge graph;
see, e.g., the blue line in Figure~\ref{fig:kg}(a).
Notice however that some of the edges in the path may be missing due to the incompleteness of the knowledge graph;
even in such circumstances, the model must ideally be able to answer path queries correctly.

\citet{guu} extended the existing KGE approaches to path QA.
For example, 
to answer a general path query $(s,r_1/\dots/r_k,o)$ with RESCAL,
a composite relation $r_1/\cdots/r_k$
is modeled
by matrix product $\mat{R}_{r_1} \cdots \mat{R}_{r_k}$,
and the score for the given query is
modeled by $\mat{e}_s\transpose\mat{R}_{r_1}\cdots\mat{R}_{r_k}\mat{e}_o$.
This formulation %
is also applicable to
DistMult and ComplEx, which use diagonal relation matrices.
In diagonalized models, however,
relation matrices are \emph{commutative},
in the sense that $\mat{R}_r\mat{R}_{r'}=\mat{R}_{r'}\mat{R}_{r}$ for any pair of relations $r, r'$.

Commutativity of relation matrices was not recognized as an issue in the past research %
because the main focus was on predicting the truth value of atomic triplets.
However, when path queries are concerned,
commutativity poses a problem.
Consider, for example,
a relation sequence
\begin{equation*}
  \text{fatherOf$^{-1}$/brotherOf/fatherOf}
\end{equation*}
and its permutation
\begin{equation*}
  \text{brotherOf/fatherOf$^{-1}$/fatherOf}.
\end{equation*}
Although these are two distinct paths (cf. Figure~\ref{fig:kg}(b, c)),
in bilinear models with commutative relation matrices,
they are represented by the same product of relation matrices,
which thereby makes the truth values of these permutated queries indistinguishable by their scores.

Drawing on the observation above,
this paper proposes a new KGE model called BlockHolE,
wherein relations are represented by block circulant matrices.
This makes relation matrices non-commutative, and thus
it does not suffer from the issues arising from commutativity,
yet in general manages to reduce the number of parameters compared with RESCAL.
It can be interpreted as a generalization of HolE and ComplEx,
and also subsumes RESCAL as an extreme case.
We report experimental results in both path and atomic QA tasks. %

\section{Notation and preliminaries}
\label{sec:notation}

\begin{table}[tb]
  \centering
  \footnotesize
  \begin{tabular}{>{$}l<{$}l}
    \toprule
    \text{Symbol}                             & Description                                                      \\
    \midrule
    \Rset, \Cset                              & sets of real/complex numbers                                     \\
    \relax[\mat{v}]_j                         & $j$th component of vector $\mat{v}$                              \\
    \relax[\mat{M}]_{jk}                      & $(j,k)$-component of matrix $\mat{M}$                            \\
    \mat{M}^{\text{T}}                        & transpose of $\mat{M}$                                           \\
    \overline{\mat{M}}                        & conjugate of $\mat{M}$                                           \\
    \odot                                     & componentwise (Hadamard) product                                 \\
    *                                         & circular convolution                                             \\
    \star                                     & circular correlation                                             \\
    \RE(x)                                    & real part of complex number $x$                                  \\
    \diag(\mat{v})                            & diagonal matrix with main diagonal  $\mat{v}$                    \\
    \text{circ}(\mat{v})                      & circulant matrix determined by $\mat{v}$                         \\
    \langle \mat{x}, \mat{y}, \mat{z} \rangle & sum of the componentwise products of $\mat{x}, \mat{y}, \mat{z}$ \\
    \mat{F}                                   & discrete Fourier matrix                                          \\
    \mathcal{E}                               & set of entities                                                  \\
    \mathcal{R}                               & set of relations                                                 \\
    \mathcal{F}                               & set of observed facts (triplets)                                 \\
    \mathcal{F}^*                             & set of ground truth facts                                        \\
    \mathcal{G}(\mathcal{F})                  & Knowledge graph induced by facts $\mathcal{F}$                   \\
    \bottomrule
  \end{tabular}
  \caption{List of symbols. See Secs.~\ref{sec:notation} and \ref{sec:kge} for detail.}
  \label{tab:nomenclature}
\end{table}

We first introduce symbols and notation used in this paper,
followed by some preliminaries on circulant matrices, circular convolution, correlation, and Fourier transform.
The summary of symbols can be found in Table~\ref{tab:nomenclature}.

Let $\Rset$ be the set of reals, and $\Cset$ be the set of complex numbers.
Let $[\mat{v}]_j$ denote the $j$th component of vector $\mat{v}$,
and let $[\mat{M}]_{j k}$ the $(j,k)$ element of matrix $\mat{M}$.
For a complex number $z$, vector $\mat{z}$, and matrix $\mat{Z}$,
let $\overline{z}$, $\overline{\mat{z}}$, and $\overline{\mat{Z}}$ denote their complex conjugate, respectively.

Let $\mat{x}$, $\mat{y}$, and $\mat{z}$ be $n$-dimensional (real or complex) vectors.
Let $\diag(\mat{x})$ denote an $n\times n$
diagonal matrix with the main diagonal components given by $\mat{x}$.
We write
$\mat{x}\odot\mat{y}$ to denote the componentwise product of $\mat{x}$ and $\mat{y}$;
i.e., $\mat{x} \odot \mat{y} = \diag (\mat{x}) \mat{y}$, or
$[\mat{x}\odot\mat{y}]_i=[\mat{x}]_i[\mat{y}]_i$, $i=1,\dots,n$.
We also write 
$\langle \mat{x},\mat{y},\mat{z}\rangle = \mat{x}\transpose \diag(\mat{y}) \mat{z} = \sum_{i=1}^n [\mat{x}]_i[\mat{y}]_i[\mat{z}]_i$.

For $n$-dimensional real vectors\footnote{%
  Generally, circular convolution, circular correlation, and circulant matrices are defined over $\Cset^n$.
  However, in this paper, it suffices to define them over $\Rset^n$.
}
$\mat{x},\mat{y}\in\Rset^{n}$,
$\mat{x}\cconv\mat{y}$ and $\mat{x}\ccorr\mat{y}$ denote \emph{circular convolution} and \emph{circular correlation},
respectively defined by
\begin{align*}
  [\mat{x}\cconv\mat{y}]_i &=\sum_{j=1}^{n}[\mat{x}]_{ i-j+1 }[\mat{y}]_j, \\
  [\mat{x}\ccorr\mat{y}]_i &=\sum_{j=1}^{n}[\mat{x}]_{ j-i+1 }[\mat{y}]_j, \quad i=1,\dots,n,
\end{align*}
where vector indices that do not fall in the range $1,\ldots, n$ must be interpreted by $0=n, -1=n-1, \dots, -n+1=1$.

For 
$n$-dimensional real vector $\mat{v}\in\Rset^{n}$, let
\begin{equation*}
  \cir(\mat{v})=
\begin{bmatrix}
[\mat{v}]_1 & [\mat{v}]_{n} & \dots & [\mat{v}]_3 & [\mat{v}]_2 \\
[\mat{v}]_2 & [\mat{v}]_1 & [\mat{v}]_{n} & & [\mat{v}]_3 \\
\vdots & [\mat{v}]_2 & [\mat{v}]_1 & \ddots & \vdots \\
[\mat{v}]_{n-1} & & \ddots & \ddots & [\mat{v}]_{n} \\
[\mat{v}]_{n} & [\mat{v}]_{n-1} & \dots & [\mat{v}]_2 & [\mat{v}]_1
\end{bmatrix}
\end{equation*}
be an operation that converts a vector $\mat{v}$ to a circulant matrix of size $n\times n$.

A circulant matrix
$\cir(\mat{v})\in\Rset^{n\times n}$
can be diagonalized
as $\cir(\mat{v}) = \FourierMatrix\inv\diag(\FourierMatrix\mat{v})\FourierMatrix$,
where $\FourierMatrix\in\Cset^{n\times n}$ is the discrete Fourier matrix of order $n$.
Also, circular convolution and correlation can be written in terms of $\cir(\cdot)$:
$\mat{x}\cconv\mat{y}=\cir(\mat{x})\mat{y}$,
and
$\mat{x}\ccorr\mat{y}=\cir(\mat{x})\transpose\mat{y}$.
It follows that
\begin{align}
  \mat{x}\cconv\mat{y} & = \FourierMatrix\inv (           \FourierMatrix\mat{x}  \odot \FourierMatrix \mat{y} ),  \label{eq:cconv-fourier} \\
  \mat{x}\ccorr\mat{y} & = \FourierMatrix\inv ( \overline{\FourierMatrix\mat{x}} \odot \FourierMatrix \mat{y} ).  \label{eq:ccorr-fourier}
\end{align}
These equations imply that
circular convolution and correlation can be computed in time $O(n\log n)$ using the fast Fourier transform (FFT).

\section{Knowledge graph embedding using bilinear maps}
\label{sec:kge}
A \emph{knowledge graph} is a labeled multigraph $(\mathcal{E}, \mathcal{R}, \mathcal{F})$,
where
$\mathcal{E}$ is the set of entities (or vertices),
$\mathcal{R}$ is the set of relation labels (or edge labels),
and $\mathcal{F}\subset
\mathcal{E}\times\mathcal{R}\times\mathcal{E}$
defines the observed instances of binary relations over entities (or labeled edges).
An item $(s, r, o) \in \mathcal{E}\times\mathcal{R}\times\mathcal{E}$
is called a \emph{triplet}, with $s$ and $o$ called its \emph{subject} and \emph{object}, respectively.
For every entity in $e\in \mathcal{E}$,
it is assumed that $\mathcal{F}$ contains at least one triplet $(s, r, o)$ with $s=e$ or $o=e$;
likewise, for every relation in $r \in \mathcal{R}$,
$\mathcal{F}$ is assumed to contain at least one triplet $(s, r, o)$.
Because $\mathcal{F}$ determines the sets $\mathcal{E}$ and $\mathcal{R}$ of entities and relations,
we write $\mathcal{G}(\mathcal{F})$ to denote the knowledge graph $(\mathcal{E}, \mathcal{R}, \mathcal{F})$ determined by $\mathcal{F}$.

Aside from observed triplets $\mathcal{F}$,
we also assume the presence of
a set $\mathcal{F}^* \subset \mathcal{E}\times\mathcal{R}\times\mathcal{E}$ of \emph{(ground truth) facts},
which is a strict superset of $\mathcal{F}$, i.e., $\mathcal{F} \subset \mathcal{F}^*$.
Thus, $\mathcal{F}^*$ is not fully observable.

\subsection{Knowledge graph completion}
\label{sec:kgc}

\emph{Knowledge graph completion} (KGC) is the task of
identifying the set of ground truth facts $\mathcal{F}^*$ from
observed facts $\mathcal{F} \subset \mathcal{F}^*$
(or equivalently, $\mathcal{G}(\mathcal{F}^*)$ from $\mathcal{G}(\mathcal{F})$).

A popular approach to KGC is to design
a \emph{scoring function} $\phi(s, r, o)$
quantifying %
how likely a triplet $(s, r, o)$ is true.
This scoring function is learned from the observed triplets $\mathcal{F}$,
in a way that it generalizes well to unobserved triplets $\mathcal{F}^* \backslash \mathcal{F}$;
i.e., the score must be high for both observed and unobserved facts, and it must be low for nonfactual triplets.

In \emph{knowledge graph embedding} (KGE)--based approaches to KGC, the scoring function $\phi(s,r,o)$ is defined in terms of the embeddings of entities and relations;
i.e., $s$, $r$, and $o$ are embedded as objects in a vector space,
and $\phi$ is defined in terms of some operations over these objects.

\subsection{Bilinear models for knowledge graph embedding}

Below, we describe some of the popular KGE models that use bilinear maps to define scoring functions. %

\subsubsection{RESCAL}
\label{sec:RESCAL}

RESCAL~\cite{rescal} provides the most general form of bilinear scoring function.
\begin{equation}
  \phi_\RESCAL(s,r,o) =\mat{e}_s\transpose\mat{R}_r\mat{e}_o,
  \label{eq:rescal}
\end{equation}
where $\mat{e}_s, \mat{e}_o\in\Rset^{n}$ are the vector embeddings of entities $s$ and $o$, respectively,
and $\mat{R}_r \in \Rset^{n\times n}$ is the matrix representing relation $r$.
Thus, $n^2$ parameters are required per relation,
which is not only a computational burden
but also the cause of overfitting during training~\cite{simple}.

\subsubsection{DistMult}
\label{sec:DistMult}

DistMult~\cite{distmult} is a model obtained by restricting the relation matrices $\mat{R}_r$ of RESCAL to diagonal;
i.e., $\mat{R}_r = \diag(\mat{w}_r)$, $\mat{w}_r \in \Rset^n$.
The scoring function is thus
\begin{align}
  \phi_\DistMult(s,r,o) & = {\mat{e}_s}\transpose\diag(\mat{w}_r)\mat{e}_o \nonumber \\
                        & = \langle\mat{w}_r,\mat{e}_s,\mat{e}_o\rangle. \label{eq:distmult}
\end{align}
Although the number of parameters is reduced considerably,
the scoring function~\eqref{eq:distmult} is symmetric with respect to the entities, i.e., $\phi_\DistMult (s,r,o) = \phi_\DistMult (o,r,s)$.
This is a severe limitation because
most real-world relations are non-symmetric.

\subsubsection{ComplEx: Complex embedding}
\label{sec:ComplEx}

The \emph{complex embedding} (ComplEx) \cite{complex}
represents entities and relations as $n$-dimensional vectors as in DistMult,
but their components are complex-valued.

The scoring function of ComplEx is given by
\begin{align*}
  \phi_\ComplEx (s,r,o) & = \RE( {\mat{e}_s}\transpose \diag(\mat{w}_r) \overline{\mat{e}_o}) \\
                        & = \RE( \langle \mat{w}_r , \mat{e}_s , \overline{\mat{e}_o} \rangle ),
\end{align*}
where $\mat{e}_s , \mat{e}_o , \mat{w}_r \in \Cset ^n $ are the embeddings of $s$, $o$, and $r$, respectively.
The number of parameters in ComplEx is $2n|\mathcal{E}|+2n|\mathcal{R}|$, and
the score is computable in time linear in the dimension of vector space.
Unlike DistMult, ComplEx can model non-symmetric relations, since $\phi(s,r,o)\neq\phi(o,r,s)$ in general.

\subsubsection{HolE: Holographic embedding}
\label{sec:HolE}

The \emph{holographic embedding} (HolE) \cite{hole} uses circular correlation to define a scoring function
\begin{equation}
  \phi_{\HolE}(s,r,o)=\mat{w}_r\transpose(\mat{e}_s\ccorr\mat{e}_o),
  \label{eq:hole}
\end{equation}
where
$\mat{w}_r,\mat{e}_s,\mat{e}_o\in\Rset^{n}$ are $n$-dimensional real vectors representing relation $r$, and entities $s$ and $o$, respectively.
HolE has only $n$ parameters per relation,
and
it can model non-symmetric relations since $\phi_{\HolE}(s,r,o)\neq\phi_{\HolE}(o,r,s)$ in general.
Computing circular correlation requires $O(n \log n)$ time if FFT is employed.
Eq.~\eqref{eq:hole} is not a bilinear form,
but it has been shown \cite{eq} that HolE is isomorphic to ComplEx,
and thus any model in HolE can be converted to an equivalent model in ComplEx, and vice versa.

\section{Path question answering over a knowledge graph}
\label{sec:path-qa}
\subsection{Path query answering}
\label{sec:path-qa-task}

Let $\mathcal{F}^*$ be the set of ground truth facts, and let $\mathcal{G}(\mathcal{F^*}) = (\mathcal{E}, \mathcal{R}, \mathcal{F}^*)$ be its induced knowledge graph.
For $k$ relations $r_1, \ldots, r_k \in \mathcal{R}$, we call
$r_1/\dots/r_k$ a \emph{relation path} of length $k$.
When $k=1$, the relation path is \emph{atomic}; otherwise, it is \emph{composite}.
Let $s, o\in \mathcal{E}$.
We say a \emph{path query} $(s, r_1/\dots/r_k, o)$ holds (or ``is true'') in $\mathcal{G}(\mathcal{F^*})$
(or with respect to $\mathcal{F}^*$)
if
\begin{multline*}
  \exists e_1, \ldots, e_{k-1}\in \mathcal{E} \;\; \forall j=1, \ldots, k \;\; \\
  (e_{j-1}, r_j, e_j) \in \mathcal{F}^*,
\end{multline*}
where $e_0 = s$  and $e_k = o$.
\emph{Path query answering} (path QA) is 
the task of predicting the truth value of path queries with respect to the unobserved set $\mathcal{F}^*$ of ground truth facts,
when its incomplete subset $\mathcal{F} \subset \mathcal{F}^*$ is only available.
In other words,
we want to predict that $(s,r_1/\dots/r_k, o)$ is true if a path from $s$ to $o$ exists in $\mathcal{G}(\mathcal{F}^*)$,
although some of the edges that constitute the path may be missing in the observed graph $\mathcal{G}(\mathcal{F})$.

For atomic path queries (i.e., those with length $k=1$),
path QA reduces to that of knowledge graph completion introduced in Section~\ref{sec:kgc}.
Thus, it is natural to address general path QA by extending the scoring function $\phi(s, r, o)$ %
of KGC methods
so that composite relation $r_1/\dots/r_k$ is allowed in place of atomic relation $r$;
i.e., by defining $ \phi(s,r_1/\dots/r_k,o) $.
Previous work \cite{guu} explored this direction,
which is also pursued in the rest of this paper. %

\subsection{Issues in existing KGE models applied to path QA}
\label{sec:path-qq-issues}

We now discuss the extension of existing bilinear KGE models to path QA.
We begin with RESCAL, which is the most general among existing bilinear models.
In RESCAL, %
if we assume $\mat{R}_r\transpose \mat{e}_s \approx \mat{e}_o$ for true triplets $(s,r,o)$,
we can model path QA as computing
\begin{multline}
  \phi_\RESCAL (s,r_1/\dots/r_k,o) \\
  = \mat{e}_s\transpose\mat{R}_{r_1}\cdots\mat{R}_{r_k}\mat{e}_o.
  \label{eq:rescal-path}
\end{multline}
As seen in this formula,
a composite relation is represented by the product of the matrices for atomic relations~\cite{guu}.

Likewise, DistMult and ComplEx can also be used for path QA,
by computing
\begin{multline*}
  \phi_\DistMult (s,r_1/\dots/r_k,o)\\
  = \mat{e}_s\transpose\diag(\mat{w}_{r_1})\cdots\diag(\mat{w}_{r_k})\mat{e}_o ,
\end{multline*}
and
\begin{multline*}
  \phi_\ComplEx (s,r_1/\dots/r_k,o) \\
  = \RE(\mat{e}_s\transpose\diag(\mat{w}_{r_1})\cdots\diag(\mat{w}_{r_k})\overline{\mat{e}_o}),
\end{multline*}
respectively.
However, because diagonal matrices are commutative,
the score of $(s,r_1/\dots/r_k,o)$
is equal to any path query in which $r_1, \cdots, r_k$ are permutated,
such as $(s,r_k/r_{k-1}/\dots/r_1,o)$.
That is, because $\phi(s,r_1/\dots/r_k,o)=\phi(s,r_k/r_{k-1}/\dots/r_1,o)$,
their truth values cannot be distinguished by the magnitude of scores.
More recent bilinear models such as ANALOGY\footnote{
  We categorize ANALOGY as a diagonal model because
  each $2\times 2$ block diagonal element of its relation matrices can be substituted by a single equivalent complex-valued component.
}~\cite{analogy}
and SimplE~\cite{simple} also represent relations by diagonal matrices,
and thus they can only model commutative relation paths.
Moreover, for SimplE,
which represents subject and object entities in different vector spaces,
it is not clear how it can be applied to path QA.

In the translation-based model %
TransE \cite{transe},
the scoring function is given by\footnote{%
  The original TransE defines a \emph{penalty} function, which gives a smaller value if a triplet is more likely to be true.
  We thus changed the sign to make it a scoring function in Eq.~\eqref{eq:transe}.}
\begin{equation}
  \label{eq:transe}
  \phi_\TransE (s, r, o) = -||\mat{e}_s+\mat{w}_{r} - \mat{e}_o||_2^2.
\end{equation}
\citet{guu} extended this function for a path query by
\begin{multline}
  \label{eq:transe-path}
  \phi_\TransE (s, r_1/\dots/r_k, ,o) \\
  = -\| \mat{e}_s+\mat{w}_{r_1}+\cdots+\mat{w}_{r_k}-\mat{e}_o\|_2^{2}.
\end{multline}
Thus, a composite relation is represented as the sum of the embedding vectors for its constituent atomic relations.
Unfortunately, Eq.~\eqref{eq:transe-path} is also invariant with the permutation of relations $r_1, \ldots, r_k$,
and their order is not respected.

\section{Knowledge graph embedding with block circulant matrices}
\label{sec:block-hole}
\subsection{BlockHolE}
\label{sec:blockhole}

In this section, we propose a bilinear KGE model suitable for path QA.
In this model, the relation matrices are non-commutative.
It thus respects the order of relations in a path query.
Further, it has a smaller number of parameters than RESCAL in general.
To be specific,
our model constrains the relation matrices to be block circulant.

A matrix is \emph{block circulant} if it can be written in the form
\begin{equation}
  \begin{bmatrix}
    \mat{W}^{(1 1)} & \cdots & \mat{W}^{(1 b)} \\
    \vdots & \ddots & \vdots \\
    \mat{W}^{(b 1)} & \cdots & \mat{W}^{(b b)}
  \end{bmatrix},
  \label{eq:block-circulant-matrix}
\end{equation}
where
each
$\mat{W}^{(i j)} = \cir(\mat{w}^{(i j)})$, $i,j=1,\ldots,b$, is a circulant matrix
determined by %
$ \mat{w}^{(i j)} \in \Rset^\blockSize $.
Thus, if the dimension of the matrix in Eq.~\eqref{eq:block-circulant-matrix} is $\ n\times n $, we have $n=bm$.
A block circulant matrix is non-commutative when $b\geq 2$;
i.e., for two block circulant matrices $\mat{A}, \mat{B} \in \Rset^{b \blockSize \times b \blockSize}$,
$\mat{A} \neq \mat{B}$,
$\mat{A}\mat{B} \neq \mat{B}\mat{A}$ in general%
.

Substituting a block circulant matrix of Eq.~\eqref{eq:block-circulant-matrix}
for matrix $\mat{R}_r$
in the bilinear scoring function (Eq.~\eqref{eq:rescal}) yields
\begin{multline}
  \phi_\BlockHolE (s,r,o)
  = \mat{e}_s\transpose \mat{R}_r \mat{e}_o \\
  \;
  =
    \overbrace{
      \!
      [\mat{e}_{s}^{(1)\mathrm{T}}\cdots\mat{e}_s^{(b)\mathrm{T}}]
      \!
    }^{\mat{e}_s\transpose}
    \overbrace{
      \!
      \begin{bmatrix}
        \mat{W}_{r}^{(1 1)}  & \!\!\! \cdots \!\!\! & \mat{W}_{r}^{(1 b)} \\
        \vdots               & \!\!\! \ddots \!\!\! & \vdots              \\
        \mat{W}_r^{(b 1)}    & \!\!\! \cdots \!\!\! & \mat{W}_{r}^{(b b)}
      \end{bmatrix}
      \!
    }^{\mat{R}_r}
    \overbrace{
      \!
      \begin{bmatrix}
        \mat{e}_{o}^{(1)} \\
        \vdots            \\
        \mat{e}_{o}^{(b)}
      \end{bmatrix}
      \!
    }^{\mat{e}_o},
  \label{eq:block-hole0}
\end{multline}
where
$ \mat{e}_s^{(i)}, \mat{e}_o^{(i)}\in\Rset^\blockSize $,
and
$ \mat{W}_r^{(i j)} = \cir(\mat{w}_r^{(i j)})\in\Rset^{\blockSize \times \blockSize}$,
$i, j = 1, \ldots, b$.
Recall that $n = b \blockSize$,
and thus
$ \mat{e}_s, \mat{e}_o \in\Rset^n $,
$ \mat{R}_r \in \Rset^{n \times n}$.
Using equalities
$\mat{x}\transpose ( \mat{y} \cconv \mat{z} ) = \mat{y}\transpose ( \mat{x} \ccorr \mat{z} )$ \cite{hole}
and 
$\cir(\mat{x}) \mat{y} = \mat{x} \cconv \mat{y}$
to rewrite Eq.~\eqref{eq:block-hole0}, we have
\begin{multline}
  \phi_\BlockHolE (s,r,o)                                                                                            \\
   = [\mat{e}_{s}^{(1)\mathrm{T}}\dots\mat{e}_s^{(b)\mathrm{T}}]
   \begin{bmatrix}
     \sum_{j=1}^{b}\mat{w}_{r}^{(1j)}\cconv\mat{e}_{o}^{(j)}                                                        \\
     \vdots                                                                                                         \\
     \sum_{j=1}^{b}\mat{w}_{r}^{(bj)}\cconv\mat{e}_{o}^{(j)}
   \end{bmatrix}                                                                                                    \\
   = \sum_{i=1}^{b}\mat{e}_s^{(i)\mathrm{T}} \left( \sum_{j=1}^{b}\mat{w}_{r}^{(i j)}\cconv\mat{e}_{o}^{(j)} \right) \\
   = \sum_{i,j=1}^{b} \mat{w}_r^{(i j)\mathrm{T}} \left( \mat{e}_{o}^{(j)}\ccorr\mat{e}_{s}^{(i)} \right)
  .
  \label{eq:block-hole1}
\end{multline}
We call this model \emph{BlockHolE},
after the fact that
it reduces to HolE when $b=1$; cf. Eq.~\eqref{eq:hole}.
Also, BlockHolE is identical to $b$-dimensional RESCAL when $ \blockSize=1 $ (or equivalently $ b = n $).

The number of parameters in BlockHolE is
$b \blockSize |\mathcal{E}|+b^2\blockSize |\mathcal{R}|$
(or $n |\mathcal{E}|+b n |\mathcal{R}|$),
and naive computation of Eq.~\eqref{eq:block-hole1} takes time
$O(b^2\blockSize \log \blockSize )$
using FFT.
However, we can make this computation faster by exploiting the duality of the Fourier transform, as shown below.

\subsection{Fast computation in complex space}

Using a similar technique used by \citeauthor{eq}~\shortcite{eq} %
to show the equivalence of ComplEx and HolE,
we can eliminate Fourier transform
to speed up the computation of BlockHolE scores.
We first rewrite Eq.~\eqref{eq:block-hole1} as follows:
\begin{multline*}
  \phi_\BlockHolE(s,r,o) \\
  = \sum_{i,j=1}^{b} \mat{w}_r^{(i j)\mathrm{T}}(\mat{e}_{o}^{(j)}\ccorr\mat{e}_{s}^{(i)})                                                                                 \\
  = \sum_{i,j=1}^{b} \mat{w}_r^{(i j)\mathrm{T}}\FourierMatrix^{-1}(\overline{\FourierMatrix\mat{e}_{o}^{(j)}}\odot\FourierMatrix\mat{e}_{s}^{(i)})                        \\
  = \frac{1}{\blockSize} \sum_{i,j=1}^{b} (\overline{\FourierMatrix\mat{w}_r^{(i j)}})^{\mathrm{T}}(\overline{\FourierMatrix\mat{e}_{o}^{(j)}}\odot\FourierMatrix\mat{e}_{s}^{(i)}) \\
  = \frac{1}{\blockSize} \sum_{i,j=1}^{b} \RE(\langle\FourierMatrix\mat{w}_r^{(i j)},\overline{\FourierMatrix\mat{e}_s^{(i)}},\FourierMatrix\mat{e}_{o}^{(j)} \rangle) ,
\end{multline*}
where $\mat{F}$ is the discrete Fourier matrix.
Here we used Eq.~\eqref{eq:cconv-fourier} to derive the second equation,
and $\mat{F}\inv = ( 1/\blockSize ) \overline{\mat{F}}\transpose$ to derive the third.
Defining complex vectors
$ { \mat{w}' }_r^{(i j)} = ( 1/\blockSize ) \FourierMatrix \mat{w}_r^{(i j)} $,
$ { \mat{e}' }_s^{(i)}   = \overline { \FourierMatrix \mat{e}_s^{(i)} } $, and
$ { \mat{e}' }_o^{(j)}   = \overline { \FourierMatrix \mat{e}_o^{(j)} } $
yields
\begin{multline}
  \label{eq:block-hole}
  \phi_\BlockHolE(s,r,o) \\
  = \sum_{i,j=1}^b \RE \left( \langle { \mat{w}' }_r^{(i j)} ,  { \mat{e}' }_s^{(i)} , \overline{ { \mat{e}' }_o^{(j)} } \rangle \right).
\end{multline}
On the basis of Eq.~\eqref{eq:block-hole},
we train $ { \mat{e}' }_k^{(i)} \in \Cset^\blockSize $ directly in complex space (i.e., the Fourier domain) instead of $ \mat{e}_k^{(j)}\in \Rset^\blockSize $
and use it as the vector embedding of entity $k$, for all $k\in \mathcal{E}$;
similarly, $ { {\mat{w}'}_r^{(i j)}} \in \Cset^\blockSize $ is directly trained in complex space to represent relation $r \in \mathcal{R}$.
The number of parameters in this model
is
$ 2 n |\mathcal{E}| + 2 b n |\mathcal{R}| $,
and Eq.~\eqref{eq:block-hole} can be computed in
$ O( b n ) $
time.
Typically, we set $ b\ll n $. %
For instance, in the experiment of Section~\ref{sec:exp},
we set $b = 2$ and $\blockSize = 50$, and thus $n = b\blockSize = 100$.
In this case, factor $b$ is negligible and the computational complexity is linear in $n$. %

\subsection{Modeling path QA}

BlockHolE can be used in path QA as follows.
First,
for any $\ell \in \mathcal{E}$ and $r\in \mathcal{R}$,
let
\begin{align*}
  {\mat{e}'}_\ell \transpose &= [{\mat{e}'}_\ell ^{(1)\mathrm{T}} \cdots {\mat{e}'}_\ell^{(b)\mathrm{T}}], \\
  \mat{W}'_r & =
               \begin{bmatrix}
                 \diag({\mat{w}'}_{r}^{(1 1)}) & \cdots & \diag({\mat{w}'}_{r}^{(1 b)}) \\
                 \vdots & \ddots & \vdots \\
                 \diag({\mat{w}'}_r^{(b 1)}) & \cdots & \diag( {\mat{w}'}_{r}^{(b b)})
               \end{bmatrix}.
\end{align*}
Then, Eq.~\eqref{eq:block-hole} can be rewritten as
\begin{equation*}
  \phi_\BlockHolE(s,r,o)
  =
  \RE\left(
    {\mat{e}'}_{s}\transpose
    \mat{W}'_r
    \overline{\mat{e}'}_{o}
  \right),
\end{equation*}
and we can compute the score of relation paths by
\begin{multline*}
  \phi_\BlockHolE (s, r_1/\dots/r_k, o) \\
  = \RE( {\mat{e}'}_s\transpose \mat{W}'_{r_1} \dots \mat{W}'_{r_k} \overline{\mat{e}'_o}).
\end{multline*}
Since $\mat{W}_{r_i}'\mat{W}_{r_j}'\neq\mat{W}_{r_j}'\mat{W}_{r_i}'$ for $b\geq 2$,
this scoring function respects the order of relations in $ r_1/\dots/r_k $.

\section{Experiments}
\label{sec:exp}
In this section,
we report the results of empirical evaluation investigating
the commutativity property of bilinear KGE models on the path QA task.
As expected, the proposed BlockHolE model, which uses non-commutative relation matrices,
outperformed commutative bilinear KGE models.

\subsection{Dataset and evaluation protocol}
\begin{table}[tb]
\centering
\small
\begin{tabular}{crrr}
\toprule
     &                 & WN11 & FB13 \\\cmidrule(lr){1-4}
     & Train           & 112,581 & 316,232 \\
Base & Valid           & 2,609 & 5,908 \\
     & Test            & 10,544 & 23,733 \\\cmidrule(lr){1-4}
     & Train           & 2,129,539 & 6,266,058 \\
\multirow{2}{*}{Path} & Valid           & 11,277 & 27,163 \\
     & Test-Deduction  & 24,749 & 77,883 \\
     & Test-Induction  & 21,828 & 31,674 \\
\bottomrule
\end{tabular}
\caption{Dataset provided by~\citet{guu}.}
\label{tab:dataset_guu}
\end{table}

\begin{table*}[tb]

\centering
\small
\begin{tabular}{lrrrrrrrrrrrr}
\toprule
                    & \multicolumn{6}{c}{WN11} & \multicolumn{6}{c}{FB13}                                                                                                                                                                                            \\\cmidrule(lr){2-7}\cmidrule(lr){8-13}
                    & \multicolumn{2}{c}{Base} & \multicolumn{2}{c}{Deduction} & \multicolumn{2}{c}{Induction} & \multicolumn{2}{c}{Base} & \multicolumn{2}{c}{Deduction} & \multicolumn{2}{c}{Induction}                                                            \\
\cmidrule(lr){2-3}\cmidrule(lr){4-5}\cmidrule(lr){6-7}\cmidrule(lr){8-9}\cmidrule(lr){10-11}\cmidrule(lr){12-13}
                    & P@10                     & MQ                            & P@10                          & MQ                       & P@10                          & MQ         & P@10       & MQ         & P@10       & MQ         & P@10       & MQ         \\\cmidrule(lr){1-13}
DistMult            & 45.6                     & 83.0                          & 33.5                          & 97.7                     & 29.6                          & 79.8       & 62.7       & 91.6       & 63.6       & 86.4       & 59.3       & 86.5       \\
ComplEx             & 60.9                     & 83.1                          & 68.7                          & 99.2                     & 46.1                          & 79.7       & 76.8       & 93.0       & 71.5       & 90.0       & 70.5       & 88.9       \\
RESCAL              & 51.8                     & 74.2                          & 43.2                          & 97.9                     & 51.2                          & 76.8       & 65.2       & 91.1       & 66.9       & 88.4       & 69.8       & 89.0       \\\cmidrule(lr){1-13}
$b=2,\blockSize=25$ & {\bf 80.9}               & {\bf 83.4}                    & {\bf 70.2}                    & {\bf 99.5}               & {\bf 54.9}                    & {\bf 81.0} & {\bf 79.2} & {\bf 93.2} & {\bf 75.0} & {\bf 91.5} & {\bf 71.3} & {\bf 90.0} \\
$b=4,\blockSize=25$ & 80.5                     & 75.6                          & 69.3                          & 99.2                     & 54.5                          & 77.4       & 76.2       & 92.1       & 72.1       & 90.5       & 70.9       & 89.5       \\
\bottomrule
\end{tabular}
\caption{Path QA ranking result: Comparing BlockHolE ($b=2, \blockSize=25$ and $b=4, \blockSize=25$) to other bilinear models.
The dimension of the embedding space for DistMult, ComplEx and RESCAL was set to $n=50$ as the result of grid search.}
\label{tab:exp-path-rank}
\end{table*}

\begin{figure}[tb]
\centering
\small
\includegraphics[width=\linewidth]{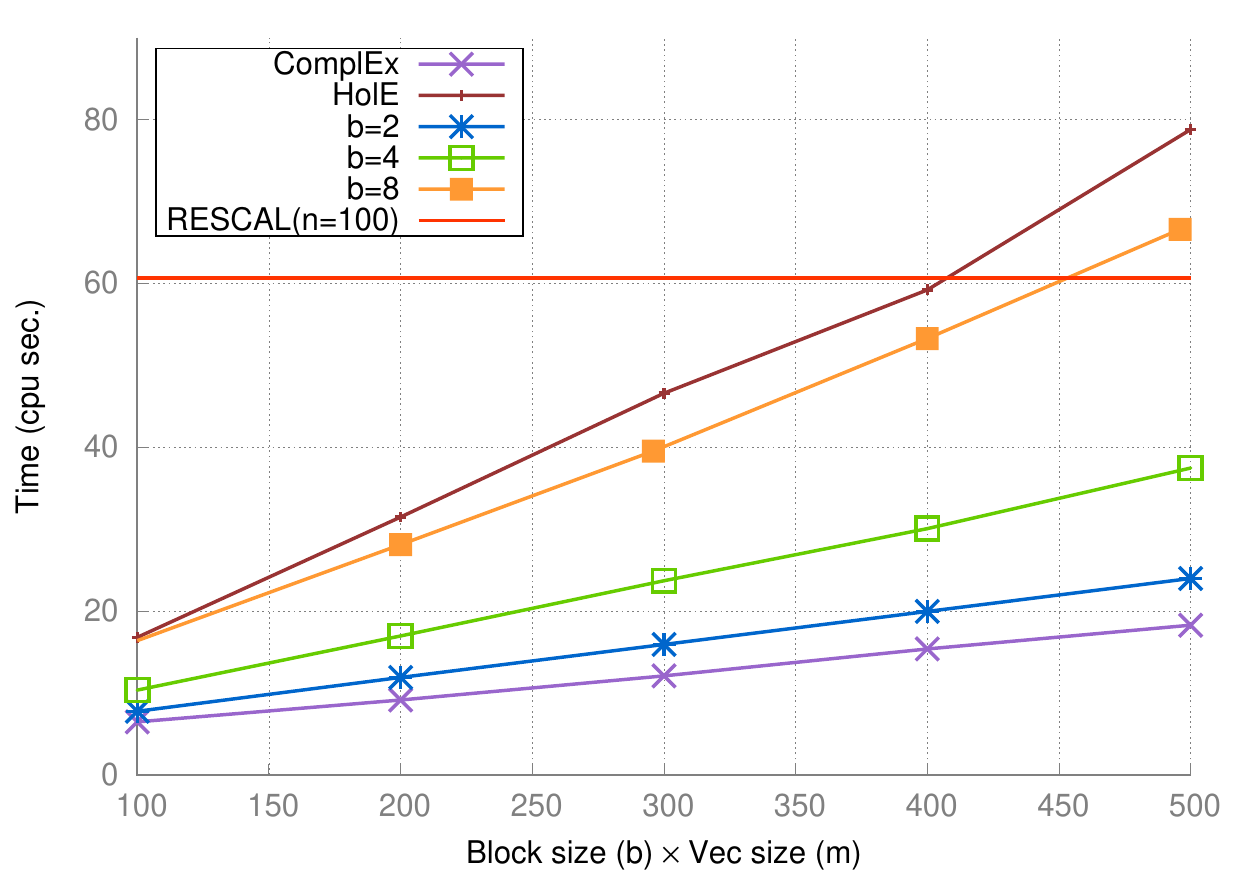}
\caption{CPU run time per epoch of BlockHolE on WN11 Base with single CPU thread.}
\label{fig:time}
\end{figure}

\begin{figure*}[tb]
\centering
\small
\begin{tabular}{c}
\begin{minipage}{0.48\hsize}
\centering
\includegraphics[width=\linewidth]{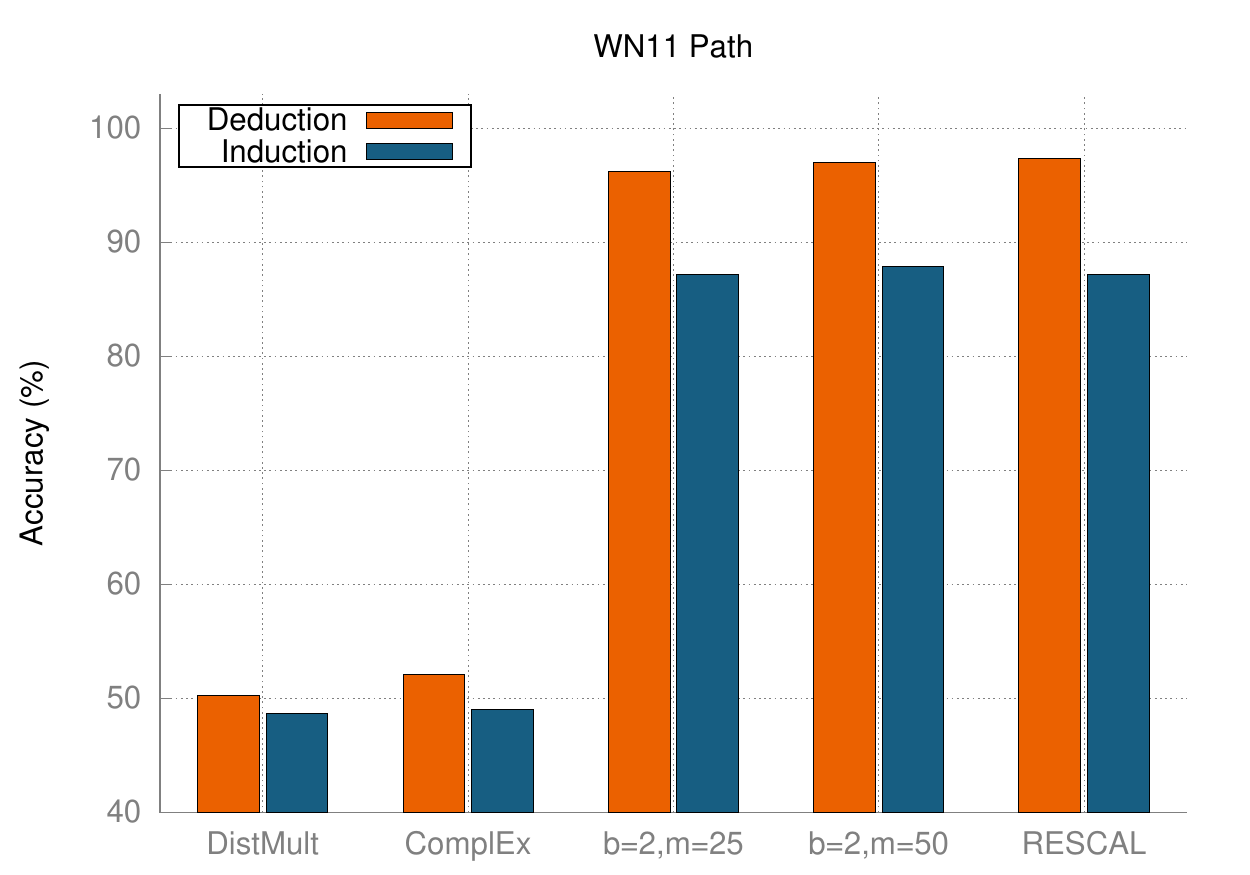}
\end{minipage}
\begin{minipage}{0.48\hsize}
\centering
\includegraphics[width=\linewidth]{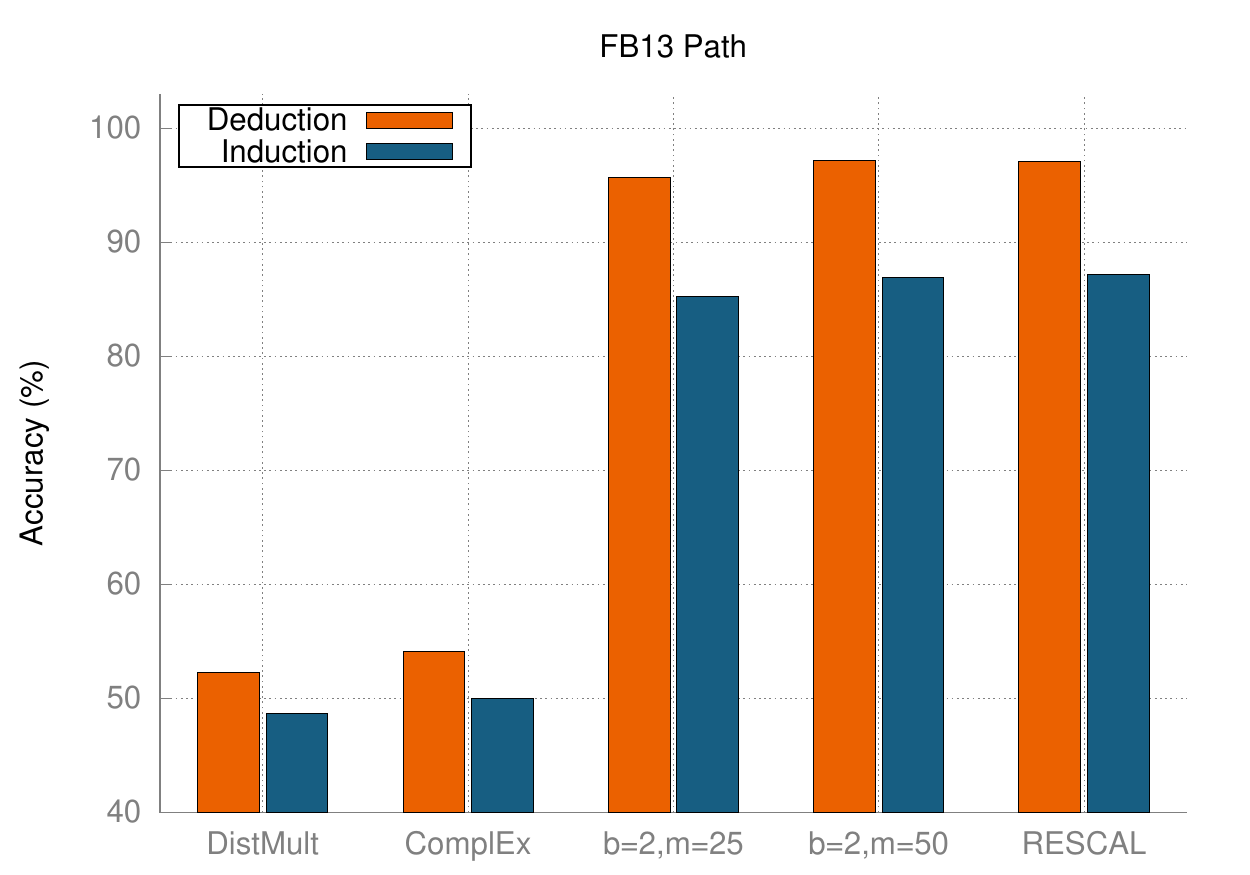}
\end{minipage}
\end{tabular}
\caption{Path QA classification result comparing BlockHolE ($b=2, \blockSize=25$ and $b=2, \blockSize=50$)
to DistMult, ComplEx and RESCAL models (all with $n=50$ as the result of grid search).}
\label{fig:acc-class}
\end{figure*}

The comparison of KGE models was performed in two path QA tasks:
(i) ranking and (ii) binary classification tasks.

\subsubsection{Path QA ranking}

For the path QA ranking task,
we adopted the same protocol and dataset used by~\citet{guu}.
Table~\ref{tab:dataset_guu} shows the statistics of their dataset.
The dataset consists of two parts, ``Base'' and ``Path''.

The Base part only contains facts (i.e., path queries with $k=1$),
and thus it is essentially for evaluating KGC performance.
Its training samples constitute the observed facts $\mathcal{F}$,
and the facts in the entire Base part (training/validation/test sets) make the ground truth facts $\mathcal{F}^*$.

The Path part contains path queries sampled from the same $\mathcal{F}$ and $\mathcal{F}^*$ as the Base part.
The test samples in the Path part is divided into ``deduction'' and ``induction'' sets.
In the ``deduction'' set,
test samples were sampled from the Base training graph $\mathcal{G}(\mathcal{F})$.
By contrast,
in the ``induction'' set, 
the test samples were chosen from the ground truth graph $\mathcal{G}(\mathcal{F}^*)$
such that none of them have a corresponding path in $\mathcal{G}(\mathcal{F})$.
Thus, the ``induction'' set is intended to measure how well a model generalizes to unobserved paths,
whereas the ``deduction'' set is to test its ability to faithfully encode the observed training graph.

At the time of evaluation,
for each a test sample $(s,r_1/\dots/r_k, \allowbreak o)$,
a \emph{candidate set} %
\begin{equation*}
\mathcal{T}(r_k)=\{t : \exists h\in \mathcal{E}\; (h,r_k,t)\in\mathcal{F}^* \},
\end{equation*}
was first computed.
In other words, the candidates are the entities for which $r_k$ (i.e., the last relation in the test query) takes
as its object at least once in $\mathcal{F}^*$.
Then, for each compared model, we made the ranking of the candidates entities in $\mathcal{T}(r_k)$
by the score $\phi(s,r_1/\dots/r_k, \allowbreak e)$,
where $\phi$ is learned by the model from the training set. %

The quality of the ranking was measured by two
evaluation metrics: averaged mean quantile (MQ)
and P@10 (percentage of correct answers ranked in the top 10).
For $q=s/p$ where $p=r_1/\dots/r_k$,
the correct answer set $\llbracket q\rrbracket$ is the set
of all entities that can be reached from $s$ by traversing
$p$ over $\mathcal{G}(\mathcal{F}^*)$.
Formally, let $\llbracket s\rrbracket=\{s\}$, and the answer set can be
recursively defined:
$\llbracket q/r\rrbracket=\{t : \exists h\in\llbracket q\rrbracket, (h,r,t)\in\mathcal{F}^* \}$.
With these definitions, MQ is computed by the following formula:
\begin{equation}
\frac{
|\{o'|o'\in\mathcal{N}(q) : \phi(s,p,o')\leq\phi(s,p,o)\}|
}
{
|\mathcal{N}(q)|
},
\label{eq:mq}
\end{equation}
where $\mathcal{N}(q)=\mathcal{T}(r_k)\setminus\llbracket q\rrbracket$ is the set of
incorrect answers.
Eq.~\eqref{eq:mq} cannot be computed for queries with which $\mathcal{T}(r_k)=\llbracket q\rrbracket$,
and these queries were excluded from evaluation.
For further details, see the original paper by~\citet{guu}.

\subsubsection{Path QA classification}

In the path QA classification task, we simply report classification accuracy.
After the scoring function $\phi$ was trained with logistic regression,
a path query $q=(s,r_1/\dots/r_k,o)$ was classified as true if $\phi(q)\geq 0$, or false otherwise.

Since the test and validation sets of Path in Table~\ref{tab:dataset_guu} contain only correct queries, we sampled negative ones by the following procedure:
For a correct query $q=(s,r_1/\dots/r_k,o)$ ($k\geq 2$),
we generated its reverse relation path query $q'=(s,r_k/r_{k-1}/\dots/r_1,o)$.
If $q'$ does not exist in $\mathcal{G}(\mathcal{F}^*)$,
we used it as a negative.

\subsection{Experiment setup}
We compared BlockHolE with state-of-the-art bilinear KGE models: DistMult, RESCAL and ComplEx.
We have implemented BlockHolE in Java. %
BlockHolE reduces to ComplEx when $b=1$,
and with the imaginary parts of parameters set to $0$,
it reduces to RESCAL when $ \blockSize=1 $ and to DistMult when $b=1$.
For a fair run time comparison, however, we separately implemented RESCAL
using jblas-1.2.4 for matrix computation.
Through all experiments, we optimized the logistic loss with
L2 regularization on the parameters $\Theta$:
\begin{equation*}
\min_{\Theta}\sum_{(q,y)\in\mathcal{D}}\log\{1+\exp(-y\phi(q;\Theta))\}+\lambda||\Theta||_2^2
\end{equation*}
where $y$ denotes the truth value of
a query $q$ in a training data $\mathcal{D}$.
Given a correct query $q=(s,r_1/\dots/r_k,o)$,
we generated negative samples by replacing $o$ with an entity randomly sampled
from $\mathcal{E}$.

We selected the hyperparameters via grid search
such that on the validation set they maximize %
classification accuracy in the path QA classification task and MQ in the path QA ranking task.
For all models except BlockHolE, all combinations of
$\lambda \in \allowbreak \{ 0.0001, \allowbreak 0 \}$,
learning rate $\eta \in \allowbreak \{ 0.005, \allowbreak 0.01, \allowbreak 0.025, \allowbreak 0.05 \}$,
and the embedding size $n\in\{50,100,150,200\}$ were tried during grid search.
For BlockHolE, 
all combinations of
$ ( b, \blockSize ) \in \allowbreak \{(2,25), \allowbreak (2,50), \allowbreak (2,100), \allowbreak (4,25), \allowbreak (4,50), \allowbreak (8,25)\}$, $\lambda \in \allowbreak \{ 0.0001, \allowbreak 0 \}$
and $\eta \in \allowbreak \{ 0.005, \allowbreak 0.01, \allowbreak 0.025, \allowbreak 0.05 \}$ were tried.
The maximum number of training epochs was set to 500.
The number of negatives generated per positive sample
was 5 during training. %

\subsection{Results}

\subsubsection{Path QA ranking}
Table~\ref{tab:exp-path-rank} shows the results on the path QA ranking data.
BlockHolE outperforms other bilinear KGE models considerably both on
deductive and inductive test settings.
These results strongly suggest that BlockHolE is more expressive in modeling path QA than DistMult and ComplEx,
while effectively reducing redundant parameters in RESCAL which can cause model overfitting.
Figure~\ref{fig:time} shows the empirical scalability of BlockHolE.
When $b$ is small, BlockHolE scales linearly in the dimension $ n = b m $ of
the embedding space.

\subsubsection{Path QA classification}
Figure~\ref{fig:acc-class} shows the accuracy of path QA classification.
DistMult and ComplEx were considerably worse than BlockHolE and RESCAL
for both WN11 and FB13.
This result confirms our claim:
The non-commutativity of relation matrices plays a critical role in modeling path QA.
The performance of BlockHolE ($b=2,\blockSize=25$) was comparable to
that of RESCAL but the former was 12 times faster.

\subsection{Analysis}
\begin{table}[tb]
\centering
\small
\begin{tabular}{ccrr}
\toprule
     Label & Relation Path & ComplEx & BlockHolE \\\cmidrule(lr){1-4}
     + & */parents/religion/* & 96.7 & 100.0 \\
     -  & */religion/parents/* & 3.3 & 100.0 \\
\bottomrule
\end{tabular}
\caption{Classification accuracy on selected queries.}
\label{tab:analysis}
\end{table}

The accuracies of BlockHolE and RESCAL on the path QA classification task
were markedly better than those of DistMult and ComplEx.
We analyzed the results further.
We extracted all queries from $\mathcal{G}(\mathcal{F}^*)$ of FB13
that consist of an interpretable relation path */parents/religion/* where $*$ denotes ``can match any relation path''.
For such queries $(s,*/\text{parents}/\text{religion}/*,o)$,
we also generated meaningless queries $(s,*/\text{religion}/\text{parents}/*,o)$ as negatives.
Table~\ref{tab:analysis} shows the classification accuracies of ComplEx and BlockHolE~($b=2,\blockSize=25$).
The results clearly show that ComplEx cannot correctly answer the negative queries at all due to the lack of the non-commutative property.

\section{Summary}
In this paper, we have pointed out the problems of existing
bilinear KGE models in path QA,
and proposed a new model that overcomes these problems.
This model, called BlockHolE, represents relations as block circulant matrices.
As a result, it respects the order of relations in path queries,
while enjoying linear-time computation of scoring functions when the number $b^2$ of blocks is sufficiently small. %
It generalizes HolE/ComplEx,
and it can also be interpreted as an interpolation between RESCAL and HolE/ComplEx.
Its effectiveness was shown empirically in path QA.%

Our proposal can be useful in not only path QA but also
many tasks such as associative rule mining~\cite{distmult}, path regularization~\cite{path_reg}, and more complex QA~\cite{dag}, in which composite relations need to be embedded as a vector.
Other future directions include reducing the increased parameters in the proposed block circulant matrices, such as by using multiplicative L1 regularization for ComplEx~\cite{l1}.

\subsection*{Acknowledgments}

We thank anonymous reviewers for helpful comments.
This work was partially supported by JSPS Kakenhi Grant Numbers 19H04173, 18K11457, and 18H03288.

\bibliographystyle{acl_natbib}

\end{document}